\documentclass{article}

\usepackage[preprint]{neurips_2025}

\usepackage[utf8]{inputenc} 
\usepackage[T1]{fontenc}    
\usepackage{hyperref}       
\usepackage{url}            
\usepackage{booktabs}       
\usepackage{amsfonts}       
\usepackage{nicefrac}       
\usepackage{microtype}      
\usepackage{xcolor}         
\usepackage{amsmath}
\usepackage{amssymb}
\usepackage{graphicx}
\usepackage{booktabs} 
\usepackage{caption} 
\usepackage{xurl}   
\usepackage{parskip} 
\usepackage{hyperref}

\usepackage{amsmath}
\usepackage{amssymb}
\usepackage{graphicx}
\usepackage{mathrsfs}

\usepackage{braket}

\newtheorem{Def}{Definition}

\newcommand{\params}{\boldsymbol{\theta}}
\newcommand{\Shannon}{\text{H}}
\newcommand{\EE}{\textbf{E}}
\newcommand{\I}{\text{I}}
\newcommand{\SSet}{\mathcal{S}}
\newcommand{\predfn}{\varepsilon}
\newcommand{\predst}{\boldsymbol{\eta}}
\newcommand{\predSSet}{\mathcal{H}}

\newcommand{\ones}{\boldsymbol{1}}
\newcommand{\stationary}{\boldsymbol{\pi}}

\title{Next-token pretraining implies 
in-context learning}

\author{%
  Paul M.~Riechers\thanks{\texttt{pmriechers@gmail.com}}
    \\
  Simplex, Astera Institute\\
\And
    Henry R.~Bigelow\\
    Simplex, Astera Institute\\
\And
    Eric A.~Alt\\
    Simplex, Astera Institute\\
\And
    Adam Shai\\
    Simplex, Astera Institute\\
}

\begin{document}

\maketitle

\begin{abstract}
We argue that in-context learning (ICL) predictably arises from standard self-supervised next-token pretraining, rather than being an exotic emergent property. This work establishes the foundational principles of this emergence by focusing on in-distribution ICL, demonstrating how models necessarily adapt to context when trained on token sequences, especially from non-ergodic sources. Our information-theoretic framework precisely predicts these in-distribution ICL dynamics (i.e., context-dependent loss reduction). We verify this with experiments using synthetic datasets of differing types of correlational structure, reproducing characteristic phenomena like phase transitions in training loss for induction head formation and power-law scaling of in-context loss. We further show that a model’s in-context performance on any task is mathematically coupled to the ensemble of tasks seen in pretraining, offering a fundamental explanation, grounded in architecture- and modality-independent principles, for such inference-time learning.
\end{abstract}

\section{Introduction}

Until about 2019,
useful AI models almost always had to be trained anew for any special purpose.  This changed drastically when it was observed that GPT-2 and its successors,
trained to minimize next-token prediction loss on a large corpus of internet data, 
could simply be prompted in context for the desired behavior---subsequently inferring and completing whatever task was requested~\cite{Radford2019_Language}.
This new regime of in-context learning (ICL)~\cite{Brown20_Language}
eventually led to the sparks of AGI that 
has 
awakened
society to a new wave of artificial intelligent systems~\cite{Bubeck23_Sparks}
with capabilities often perceived as almost magical in their ability to adapt. However, we contend that the capacity for ICL in these models is not an exotic emergent property, but rather a mathematically inevitable consequence of a model's successful minimization of the standard next-token prediction loss.
In this work, we provide a rigorous
demonstration of this implication, establishing how pretraining on token sequences \textit{necessarily} gives
rise to ICL capabilities by focusing our analysis on ICL within the statistical landscape of the training
data itself. While a full account of out-of-distribution ICL is beyond our current scope, the principles established here for in-distribution behavior offer a crucial theoretical grounding and suggest key mechanisms that may underpin more general forms of context-dependent learning.

Despite the established state of affairs, the \emph{reason} that ICL emerges so broadly in large language models (LLMs) has not been fully established.
Many recent papers have claimed that ICL mimics Bayesian inference~\cite{Xie21_Explanation, Arora24_Bayesian}, or that it mimics gradient descent~\cite{Oswald23_Transformers, Ahn23_Transformers},
however a causal narrative of which conditions are needed to attain ICL appears to be lacking.  
Here, we show both mathematically and empirically that 
ICL, including the general ability of AI models to home in on their users and novel requests~\cite{Lampinen24_Broader}, is largely 
explained as an inevitable consequences of performing well at next-token prediction.
We formalize ICL as fundamentally a process of \textit{synchronization} to the hidden structure of the training data---a natural outcome of predicting the next token effectively.

Our core results demonstrating this necessary emergence of in-distribution ICL are independent of specific ML architecture; Transformers, RNNs, or any other architecture optimized for next-token prediction should display these key features. While the specifics of how different architectures implement this capability may influence their generalization to out-of-distribution scenarios (a crucial open question), the fundamental capacity for in-context adaptation arises from the predictive objective itself. It has recently been established that various architectures, when pretrained on next-token prediction, represent belief-state geometry in their neural activations~\cite{Shai24_Transformers, Piotrowski25_Constrained, Pepper24_RNNs}. This reinforces 
the idea that models are trained to \emph{predict} rather than merely \emph{generate} data. Indeed, to predict the future as well as possible, a model benefits from maintaining an implicit generative model of the world and performing updates over its latent states as more context is observed~\cite{Shai24_Transformers, Shal98a}.


These internal belief states, akin to Bayesian updating and reducing uncertainty about a hidden world model, are sufficient to explain aspects of in-context learning. However, the mathematical necessity of ICL, as we will show, follows even more simply from the imperative of minimizing the standard next-token cross-entropy loss. Successfully doing so implies a reduction in predictive uncertainty as more context is processed. This context-dependent reduction in loss, which we can precisely anticipate for certain correlated stochastic processes, is the essence of ICL. Furthermore, the in-context loss on a particular task or pattern is not solely determined by that pattern but is influenced by the entire ensemble of tasks and patterns encountered during pretraining. The characteristics of non-ergodic data, for instance, provide a minimal model for understanding phenomena like the power-law decay often observed in ICL performance and can elucidate the utility of architectural features like induction heads~\cite{Elhage21_Mathematical, Olsson22_Context} as mechanisms for disambiguating latent data sources. Indeed, in this work, we create and verify toy models that demonstrate these core phenomena.


\paragraph{Our primary contributions are:}
\begin{itemize}
    \item We establish that ICL is a mathematically inevitable consequence of optimizing standard next-token prediction objectives. We demonstrate this foundational principle for in-distribution ICL, particularly showing its richness when training on non-ergodic data.
    \item We provide a framework to precisely anticipate ICL dynamics for models trained on correlated stochastic processes, verifying this with toy models where ground truth is known.
    \item We theoretically predict and validate how a model's in-distribution ICL performance on a specific pattern or task is mathematically determined by, and coupled with, the entire ensemble of patterns encountered during its training.
\end{itemize}


\section{ICL from pretraining on any architecture}

A sequence of tokens $x_{1:L} \in \mathcal{X}^L$ can be thought of as a realization of a sequence of correlated random variables $X_{1:L}$.
A neural network's
weights $\params$ induce a parametrized conditional probability distribution $\Pr_{\params}(X_{\ell+1} | X_{1:\ell})$ over the next token 
for each
preceding context.
If training sequences of 
length $L$ are sampled from a distribution
$Q(X_{1:L})$,
then the expectation of population loss $\braket{\mathcal{L}_{\params}}$ is the cross entropy, which can be expressed as an accumulation of terms over the context window:
\begin{align}
	\braket{\mathcal{L}_{\params}}_{Q(X_{1:L})} 
	&= - \braket{ \log \Pr_{\params} (X_{1:L})}_{\! Q(X_{1:L})} 
    \!
    =
    \sum_{\ell = 1}^L - \braket{ \log \Pr_{\params} (X_\ell | X_{1:\ell-1})}_{\! Q(X_{1:\ell})}  
    \!
	=  \sum_{\ell = 1}^L c_\ell^{(\Pr_{\params}, Q)}
    . 
\end{align}	
Each term in this sum, $c_\ell^{(\Pr_{\params}, Q)} := - \braket{ \log \Pr_{\params} (X_\ell | X_{1:\ell-1})}_{Q(X_{1:\ell})}$, is the \emph{myopic cross-entropy rate}—the expected loss specifically for predicting the token at position $\ell$ given the context $X_{1:\ell-1}$.

If some setting of the parameters very well approximates the true distributions such that $\Pr_{\params}(X_{1:L}) = Q(X_{1:L})$,
then 
the expected loss will be minimized by the joint entropy $\Shannon[Q(X_{1:L})]$:
\begin{align}
	\min_{\params}
	\braket{\mathcal{L}_{\params}}_{Q(X_{1:L})}  
	&= 
    - \braket{ \log Q(X_{1:L})}_{Q(X_{1:L})} 
    =
    \sum_{\ell = 1}^L - \braket{ \log Q(X_\ell| X_{1:\ell-1})}_{Q(X_{1:\ell})} 
	=  \sum_{\ell = 1}^L  h_\ell^Q
	~.
\end{align}

Note that the minimized loss is the accumulation of the \emph{myopic entropy rate}
$h_\ell^Q := - \braket{ \log Q(X_\ell| X_{1:\ell-1})}_{Q(X_{1:\ell})} $,
which lower bounds the myopic cross-entropy rate 
\begin{align}
c_\ell^{(\Pr_{\params}, Q)}  & \geq h_\ell^Q  \quad \text{ at each } \ell
\end{align}	
since 
$-\braket{ \log \Pr_{\params}(X_\ell|  x_{1:\ell-1})}_{Q(X_\ell|  x_{1:\ell-1})}  \geq - \braket{ \log Q(X_\ell|  x_{1:\ell-1})}_{Q(X_\ell| x_{1:\ell-1})} $ 
%
for each specific context $x_{1:\ell-1}$,
with equality if and only if $\Pr_{\params}(X_\ell|  x_{1:\ell-1}) = Q(X_\ell|  x_{1:\ell-1})$.
If expected loss is minimized at $\params^*$,
then 
$c_\ell^{(\Pr_{\params^*}, Q)}  = h_\ell^Q = h_\ell^{\Pr_{\params^*}}$ 
at each $\ell$---i.e., the myopic cross-entropy rate asymptotes to the myopic entropy rate as training converges, at each position in the context window for transformer-type models, and at each timestep for RNN-type models.
For these well-trained models, 
their uncertainty over the next token
$h_\ell^{\Pr_{\params^*}}$
will generically reduce as more context is taken into account with larger $\ell$.

For stationary processes,
$h_{\ell+1}^Q \leq h_\ell^Q$, so \emph{the next-token entropy of predictions from 
	any model well pretrained on an approximately stationary process must go down as it sees more context}.






\section{Anticipating exact in-context entropy convergence}

In-context loss, as a function of context position, approaches the myopic entropy rate during pretraining.
This context-dependent reduction in 
entropy of predictions 
can be understood and calculated exactly 
via the context-induced distributions over latent states of a generative world model~\cite{Riec18_SSAC1}.
In particular, it can be useful to consider an explicit construction of the observation induced metadynamic among all belief states---the so-called `mixed-state presentation' (MSP).
The myopic entropy rate can be calculated exactly via linear algebra and non-normal spectral theory 
using the net stochastic transition operator over these belief states~\cite{Riec18_SSAC2}.

\subsection{Metadynamic among predictive states}
\label{sec:PredictiveMetadynamics}

Let's say that you've already learned what the world is like---all of its deterministic features, random elements, and how bits of randomness feed back into the process to determine future behavior.
Now you want to leverage this knowledge to iteratively predict the immediate future as well as possible.  I.e., you want to track conditional probability distributions $Q(X_{L+1} | X_{1:L} = w)$
over the next token as you observe the specific realization of history.
Naively, this may seem to require storing a belief for each distinct past---but the number of possible pasts grows exponentially with the length of the sequence.
For any agent with finite memory, this explosion of memory states would severely curtail their ability to predict.

However, not all histories need to be distinguished:
Instead, we can track conditional densities over the full future (rather than just the next token),
and cluster those histories that yield the same conditional future distribution,
providing the minimal sufficient statistic for 
iteratively predicting the next-token distribution~\cite[Corollary 2]{Shal98a}.
This significantly reduces the resources required to predict.

In other words,
for the purpose of prediction, we 
can cluster those histories that yield the same 
full-future predictions,
as done routinely in computational mechanics~\citep{Uppe97a, Shal98a,  Crut12a, Riec18_SSAC1}.
To formalize this simple idea, we consider
the 
equivalence classes
of 
arbitrary-length histories $w \in \mathcal{X}^*$
that induce the same conditional probability distributions over futures.
\begin{Def}
	Two histories 
	$w$ and $w'$
	(of arbitrary lengths 
	$\ell$ and $\ell'$)
	belong to the same 
	\emph{predictive state}
	$\predfn(w) = \predfn(w')$
	if 
	$ Q( X_{\ell+1:\ell+k} | X_{1:\ell} = w) = 
	Q( X_{\ell'+1:\ell'+k} | X_{1:\ell'} = w')$ 
	for all $k \geq 1$.
\end{Def}
These predictive states
are the \emph{minimal set of maximally predictive features}.
The set of 
predictive states
$\predSSet$
partitions the set of all histories $\mathcal{X}^*$.
The 
\emph{predictive-equivalence function} 
$\predfn \colon  \mathcal{X}^* \to \predSSet$
maps an observed history of any length to 
its 
predictive state.
The empty history $w =  \varnothing$, when no observations have yet been made, belongs to the 
predictive state
$\predfn(\varnothing) = \bigr\{ w \in \mathcal{X}^* :Q( X_{|w|+1:|w|+k} | X_{1:|w|} = w) = Q( X_{1:k} )$ 
for all $k \geq 1 \bigr\}$.

Note that 
the current predictive state $\predst$ and the next token $x$ uniquely determine the next predictive state $\predst'$.
During inference, 
the apparent dynamic over tokens
thus induces a metadynamic over predictive states (i.e., a metadynamic over  minimal sufficient statistics of historical context).~\footnote{In contrast, next-token distributions are insufficient memories for recurrent prediction, since two distinct predictive states can have identical next-token distributions and merging these predictive states would cause worse next-token prediction at some point in the future.}

It is useful to consider a vector space where the distinct predictive states form an orthonormal basis.
There is then a linear stochastic transition operator $W$ among predictive states with matrix elements $W_{\predfn(w) \to \predfn(wx)} = \Pr(x | \predfn(w))$.
There is a linear functional $\ket{\Shannon(X|H)} := \sum_{\predst \in \predSSet} \ket{\delta_{\predst}} \Shannon(X|H = \predst) $ on this space that corresponds to the next-token entropy from each predictive state~\footnote{Following the tradition of the Markov-chain literature, we choose state distributions to be row vectors and linear functionals on this space to be column vectors.}.
The myopic entropy rate can then be anticipated in closed form via $h_\ell^Q = \braket{\delta_{\predfn(\varnothing)} | W^{\ell-1} | \Shannon(X|H)}$.

This general construction is valid for any stochastic process for correlated tokens.  To verify that this construction precisely predicts in-context learning in neural networks, 
we construct some toy scenarios 
where this claim could be falsified.
In the next section we will see how to construct this predictive metadynamic from any HMM generator of the token sequences.
We will see that in-context learning indeed proceeds exactly as anticipated in these scenarios.



\subsection{The MSP of an HMM}

To make quantitative predictions about 
in-context learning abilities of neural networks, we will consider synthetic training data from hidden Markov models (HMMs).
Allowing for infinitely many hidden states makes HMMs as expressive as any class of data generators, transcending the limitation to regular languagues that afflicts finite-state HMMs.
We also allow our HMMs to have multiple ergodic components.

We employ Mealy HMMs of the form $\mathcal{M} = (\mathcal{X}, \SSet,\predst_\varnothing, (T^{(x)})_{x \in \mathcal{X}})$,
where 
$\mathcal{X}$ is the token alphabet,
$\SSet$ is the set of latent states,
$\predst_\varnothing$ is the initial distribution over latent states,
and each observable token has a corresponding substochastic transition matrix 
$T^{(x)}$ whose matrix elements 
$T_{s,s'}^{(x)} = \Pr(x,s' | s)$
encode the conditional joint probability of next observing token $x$ and transitioning to latent state $s'$,
given that the generator was in latent state $s$.

The substochastic transition matrices sum to a row-stochastic matrix $T = \sum_{x \in \mathcal{X}} T^{(x)}$.
Note that $T$ has a stationary right eigenvector $\ones = T \ones$
where $\ones = [1 \, 1 \, \dots 1]^\top$
is the length-$|\SSet|$ column vector of all ones.
The stationary left eigenvector $\stationary = \stationary T $, 
normalized such that $\stationary \ones = 1$,
corresponds to the stationary distribution over latent states and is generically non-uniform.
If $|\SSet| < \infty$, then a stationary process can always be attained by setting $\predst_\varnothing = \stationary$.

The probability of any sequence $w = x_1 x_2 \dots x_\ell$ can be calculated from its HMM via linear algebra since 
$Q(X_{1:\ell}=w) = \predst_\varnothing T^{(w)} \ones$,
where $T^{(x_1 x_2 \dots x_\ell)} = T^{(x_1)} T^{(x_2)} \cdots T^{(x_\ell)}  $.
As always, conditional probabilities can be calculated by the joint divided by the marginal:
\begin{align}
Q(X_{\ell+1:\ell + \ell'} | X_{1:\ell} = w) = 
\frac{\predst_\varnothing T^{(w)}}{\predst_\varnothing T^{(w)} \ones} T^{(X_{\ell+1:\ell + \ell'})} \ones
~.
\end{align}
From this, we note that any two sequences $w$ and $w'$ that induce the same \emph{predictive vector} (i.e., the vector representation of the distribution over latent states $\frac{\predst_\varnothing T^{(w)}}{\predst_\varnothing T^{(w)} \ones} = \frac{\predst_\varnothing T^{(w')}}{\predst_\varnothing T^{(w')} \ones} $), must induce the same conditional probability density over all possible future sequences---i.e., these different contexts induce the same predictive state, $\predfn(w) = \predfn(w')$.
Moreover, if two contexts $w$ and $w'$
induce nearby predictive vectors, then
they cannot significantly differ in how they affect the future.

For any HMM, the predictive metadynamic
can be constructed from the observation-induced transitions among distributions over latent states, since each distribution over latent states implies a probability density over all future token sequences.  
The result of this construction is typically called the `mixed-state presentation' (MSP)~\cite{Crut16_Exact, Jurgens21_Shannon},
since predictive vectors are typically called `mixed states'~\cite{Uppe97a}.  
This is a dynamic over belief-states---a dynamic over latent distributions of the generative model---and can thus be thought of as a predictive metadynamic for optimal inference as the token dynamic is observed in sequence.
The MSP is itself an HMM, with a unique initial state delta distributed on the initial distribution over latent states of the original generative HMM:
$\mathcal{M}_\text{MSP} = (\mathcal{X}, \mathcal{R}, \bra{\delta_{\predst_\varnothing}}, (W^{(x)})_{x \in \mathcal{X}})$.

\begin{figure}
\centering
	\includegraphics[width=0.99\textwidth]{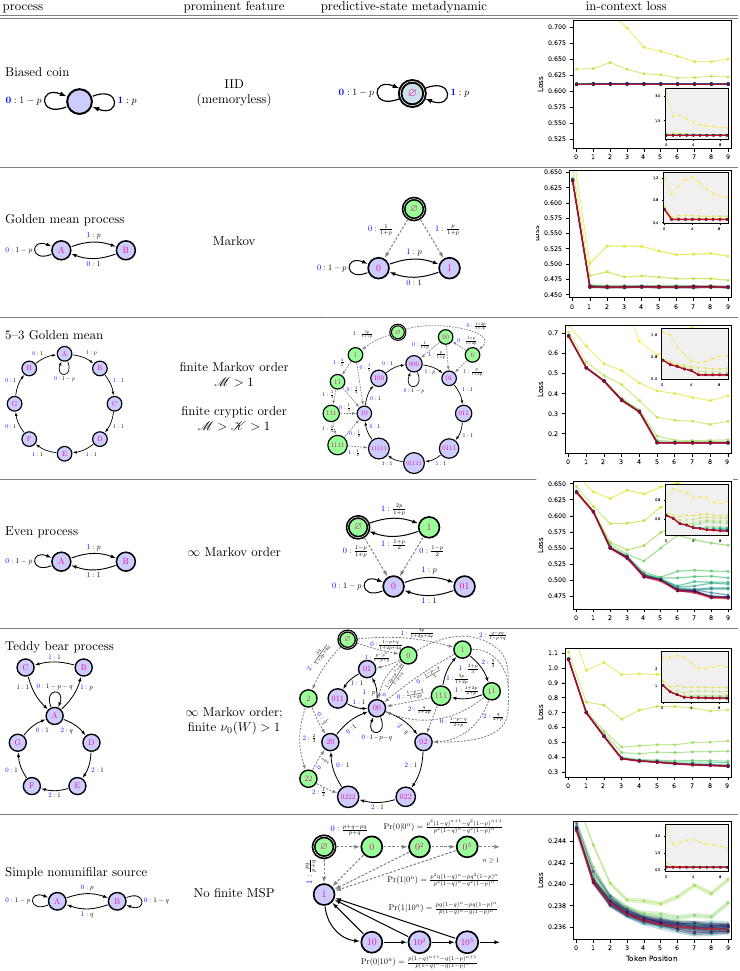}
	\caption{
    Simple processes producing tokens with correlations of increasing complexity.  In-context learning, the reduction of predictive uncertainty with increasing context,
    is a signature of the data's correlational structure.
    The third column shows the MSP induced by updating belief states, given knowledge of the data-generating HMM in the first column.
    The states of the MSP are labeled by the shortest context that induces them.  Transient states of the MSP are green, while recurrent belief states are purple.  
    We highlight the unique initial state $\predst_\varnothing$ of the MSP by a concentric circle.
    Thick red line in the right-most column shows the theoretical expectation of the myopic entropy rate---the in-context reduction of uncertainty---as a function of context position.  
    Thin lines 
    show in-context loss as transformers converge to this theoretical curve, with line color progressing from light to dark as training progresses. Insets are zoomed out, showing loss for all checkpoints.
	}
\label{fig:MyopicEntropy1}
\end{figure}

The set of mixed states 
induced by observable sequences is:
$\mathcal{R} = \bigl\{ \frac{\predst_\varnothing T^{(w)}}{\predst_\varnothing T^{(w)} \ones} : \, w \in \mathcal{X}^*, \,  \predst_\varnothing T^{(w)} \ones > 0\bigr\}$.  Each element $\predst$ of $\mathcal{R}$ can be encoded as a one-hot column vector $\ket{\delta_{\predst}}$
or a one-hot row vector 
$\bra{\delta_{\predst}}$,
such that 
$\braket{\delta_{\predst} | \delta_{\predst'}} = \delta_{\eta, \eta'}$.
The elements of the MSP's substochastic transition matrices are 
\begin{align}
W_{\predst, \predst'}^{(x)} = \braket{\delta_{\predst} | W^{(x)} |\delta_{\predst'}} = \Pr(x, \predst' | \predst) = \eta T^{(x)} \ones \, \delta_{\predst', \frac{\eta T^{(x)} }{\eta T^{(x)} \ones }} ~.
\end{align}
The net stochastic transition operator among mixed states,
$W = \sum_{x \in \mathcal{X}} W^{(x)}$
serves precisely as the transition operator among predictive states,
which we discussed above for general processes of correlated token sequences.~\footnote{The main nuance is that the MSP of a non-minimal HMM may retain redundancies, such that two mixed states belong to the same predictive state.  Constructing the MSP of the MSP will remove such redundancies.}

\subsection{Anticipating exact in-context entropy convergence in ergodic toy models}

To 
visualize
the mixed-state presentation, and its implications for ICL, we start simply with ergodic stochastic processes that can be generated by finite 
HMMs with a single ergodic component.
Despite the simplicity of this setting, 
Fig.~\ref{fig:MyopicEntropy1} shows 
increasing levels of nuance in the correlational structure of these stochastic processes, which 
neural networks must address in context 
even after it has learned the generative mechanism for the data.
The directed edges of the automata, from latent state $s$ to $s'$, 
are labeled as ``${\color{blue}x}: \Pr(x,s'|s)$'' to indicate the probability of observing token $x$ and moving to latent state $s'$ from latent state $s$.
When we train transformers on sequences generated from these processes, we observe the in-context loss converge to the myopic entropy rate over the course of training.  
All experimental curves are plotted with error bars of $\pm 1$ standard error of the mean.
Stepping through the rows of Fig.~\ref{fig:MyopicEntropy1} 
highlights
how distinct correlational features of training data affect in-context loss.

For an uncorrelated process, exemplified by a biased coin but true of any IID process,
historical context doesn't inform the future, and there is thus no in-context learning.
Markovian processes, exemplified by our Golden Mean process that cannot produce consecutive `1' tokens, hit the asymptotic in-context loss after a single observation.
Non-Markovianity then is the key feature of non-trivial in-context learning.
The \emph{Markov order} of a process $\mathscr{M} = \min(\{ \ell \in \mathbb{N}_{\geq 0} : Q(X_{\ell':\infty} | X_{\ell'-\ell:\ell'-1}) = Q(X_{\ell':\infty} | X_{1:\ell'-1}) \text{ for all } \ell'> \ell \})$ is the minimal distance into the past one needs to look to perform perfect inference.
IID processes are Markov-order 0, 
while Markov processes are Markov-order 1.
If a process has finite Markov order,
like our 5-3 Golden Mean process of Markov-order 5, then 
in-context loss hits the asymptotic value after the $\mathscr{M}^\text{th}$ token is observed in a sequence. 
After $\mathscr{M}$ tokens, 
all probability density has flushed out of the transient states of the MSP.

\begin{figure}
\centering
	\includegraphics[width=0.99\textwidth]{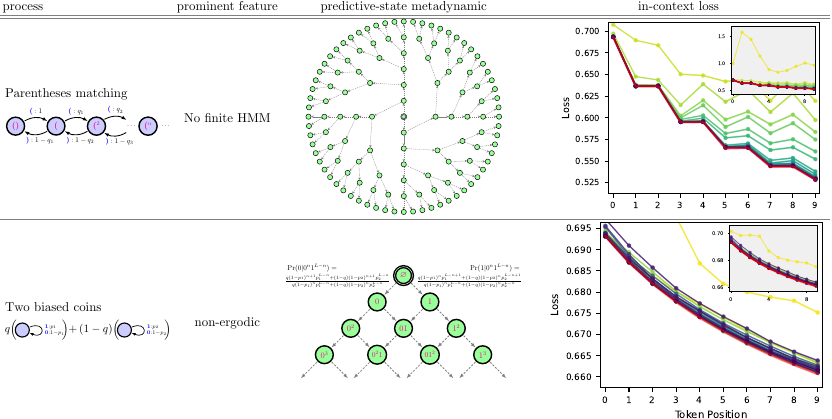}
	
	\caption{
    Simple processes beyond the paradigm of finite ergodic generators.  As with the processes in the previous table, predictive uncertainty decreases with increasing context in accordance with the predictive metadynamic.
    On the right, we see transformers' in-context loss converging to our theoretical expectation (thick red line) over the course of training (light to dark lines).
	}
\label{fig:MyopicEntropy2}
\end{figure}

However, even simple models with a small number of latent states can have infinite Markov order.
The Even process, which must have an even number of `1' tokens between any two `0' tokens, has only two latent states but infinite Markov order.  
In this case, probability density decays exponentially out of the MSP's transient states of uncertainty, with a corresponding exponential decay of in-context loss.
Correlated processes typically contain coexisting mechanisms for resolving ambiguity, depending on the particular sequence observed.
Our Teddy-bear process shows an ephemeral dropout from certain transient states of the MSP within finite time,
while other transient states of uncertainty may linger 
for certain arbitrarily long sequences.
We note that certain ambiguities must be resolved before others, as exemplified by the transient MSP structure of the RRXOR process of Ref.~\cite[Fig.~9]{Riec18_SSAC2}.
Finally we note that simple motifs in the transition dynamic among latent states imply infinitely many predictive states, as in the Simple Nonunifilar Source in the last row of Fig.~\ref{fig:MyopicEntropy1}.

These nuanced features of in-context resolution of ambiguity have all
been within the realm of correlated processes generated by finite ergodic HMMs.
Moving either (i) up the Chomsky hierarchy (from regular to 
context-free, context-sensitive, or unrestricted grammars)~\cite{Lin17_Critical} or (ii) into the realm of non-ergodic processes~\cite{Crut15a} is sufficient to induce power-law decay of in-context loss.
The transient structures of the corresponding predictive metadynamics, and the corresponding in-context loss for transformers trained on these processes, are shown in Fig.~\ref{fig:MyopicEntropy2} for simple examples of context-free (Parentheses Matching) and non-ergodic (Two Biased Coins) processes.



    
\section{Induction heads disambiguate ergodic components with distinct support}
\label{sec:InductionHeads}

One of the major insights from analyzing the mathematical structure of transformers is that the attention mechanism easily allows for induction heads that successfully predict skip trigrams of the form [A][B]$\dots$[A] $\mapsto$ [B]~\cite{Elhage21_Mathematical, Olsson22_Context}.
Memorably, the Anthropic interpretability team
showed that such induction-head circuits 
predict `Dursley' after `Mr.' if `Mr.\ Dursley' had appeared earlier in the text~\cite{Elhage21_Mathematical}.
But when is such behavior useful?  Such behavior is especially useful when an alternative text---generated from a distinct ergodic structure---would have been talking about Mr.\ So-and-so instead.  I.e., induction heads are useful when the training data is sampled from a nonergodic source, which is indeed very much the case when scraping an internet's worth of data.  In addition to Harry Potter books featuring Mr.\ Dursley, there are Charlie and the Chocolate Factory books featuring Mr.~Wonka and, of course, countless other books and themes to disambiguate.

\begin{figure}[t]
\centering
	\includegraphics[width=1.0\linewidth]{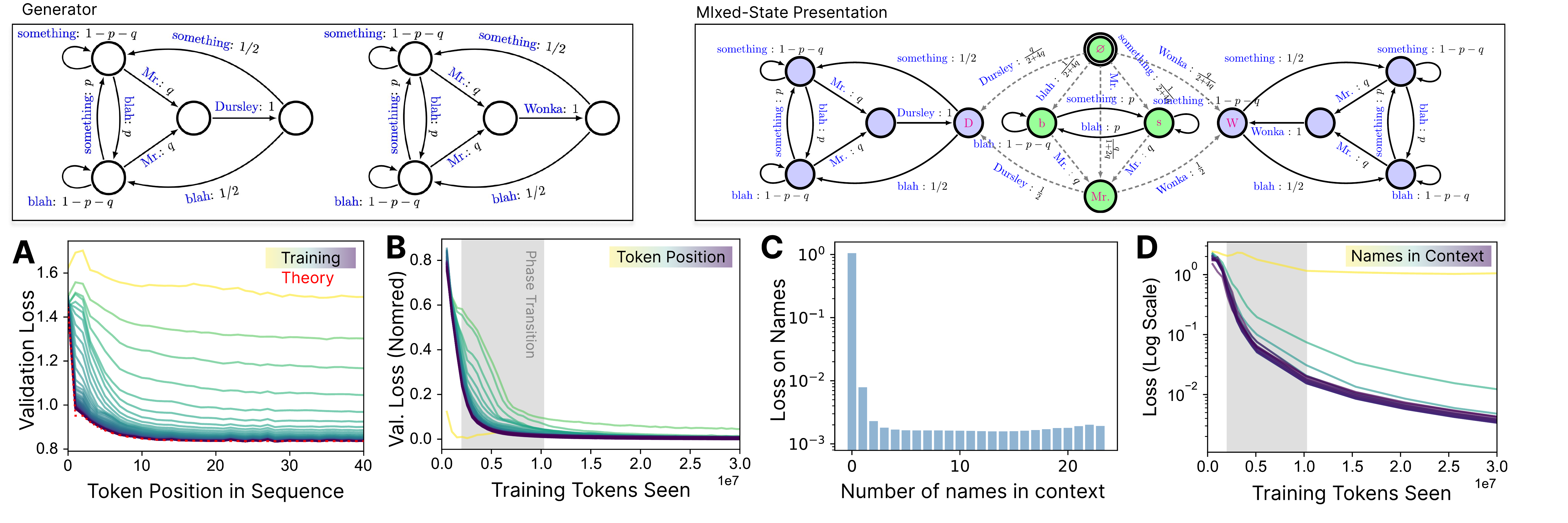}
	
	\caption{
		The need for induction heads can easily be understood via nonergodic processes.  In this simple example, induction heads allow the two disconnected components---one about Harry Potter and the other about Willy Wonka---to be distinguished.  
		Example sequence at $p=2/5$ and $q=1/4$, with 50\% probability on each ergodic component:
        ``{\color{blue} something something something Mr. Wonka blah blah blah something something something blah something blah Mr. Wonka blah something something} $\dots$''
        \textbf{A}) Validation loss decreases not only over training (light to dark lines) but also over token position (x-axis) as context from earlier tokens makes the prediction of subsequent tokens easier. 
        Increased training enables the models to more effectively make use of this context. 
        \textbf{B}) 
        Plotting loss over training for each token position (lighter lines correspond to earlier token positions) 
        reveals the characteristic ``bump'' of the ``phase change'' associated with the development of induction heads~\cite{Olsson22_Context}. \textbf{C}) 
        Loss of a trained model when it predicts the name tokens (Wonka/Dursley) in a sequence, as a function of how many times a name previously occurred in context.  This reveals the model's initial uncertainty before it has encountered a name, and its subsequent high confidence in predicting the right name. \textbf{D}) This ability to confidently predict the correct name, conditioned on seeing instances of the name earlier in the context, is learned over the course of training.
        %
	}
\label{fig:InductionHeads}
\end{figure}

Fig.~\ref{fig:InductionHeads} shows a toy model that captures the essence of the situation.
Two different ergodic components---one about Mr.~Dursley and one about Mr.~Wonka---contribute to the training data.  
In context, induction heads allow disambiguating which ergodic component the sequence came from, and thus reduces the myopic entropy after the first encounter with either Dursley or Wonka.
Similar in-context symmetry-breaking mechanisms
will abound as predictive models refine their understanding of where they should be in idea space.


%
%

\section{Non-ergodicity and hierarchical ambiguity}

A natural hierarchy of ambiguity in written text includes increasingly refined resolution of language, genre, author, and attitude.  Moving down this hierarchy corresponds to symmetry breakings of knowledge, as an LLM homes in on the distinguishing characteristics of its user~\cite{EggSyntax24_Language}.

There are varying levels of nuance in this disambiguation.
If two ergodic components have a different token alphabet, as in the Wonka--Dursley example above, then 
in-context uncertainty drops suddenly, with probability density flushing entirely out of certain parts of belief space, when certain tokens are observed.  
However, even if two ergodic components have the same alphabet, they can have different languages---different token sequences that are forbidden in that component.
Yet more nuanced, different ergodic components can have the same alphabet and same language, but may merely differ in the probability distribution over token sequences, as 
with the Two Biased Coins example.

In practice, many ergodic components will have a mixture of these nuanced differences.  This indicates different timescales of disambiguation via different mechanisms.  During inference, an ideal predictor may 
come to rule out certain ergodic components instantaneously, over time, or only asymptotically, depending on the details of the script under scrutiny.

It is known that training on 
non-ergodic processes generically yields a power-law decay of myopic entropy when the parameters of an ergodic component are sampled from a continuous probability density for each sequence~\cite{Crut15a, Bialek01_Predictability}. 
%
%
Power-law scaling of myopic entropy can independently arise from the 
nested correlations
of context-free and context-sensitive grammars, even for an ergodic process~\cite{Lin17_Critical}.  
Natural language incorporates both of these features and so in-context power-law behavior has at least two distinct mechanistic origins.


\subsection{In-context learning as a function of the number of ergodic components}

What is the nature of in-context loss when neural networks are trained on nonergodic processes?  This is the practically very relevant case where the statistics within a typical sequence are not representative of the statistics over all sequences.


Any nonergodic process with $N$ ergodic components
has the joint distribution 
\begin{align}
Q_N(X_{1:\ell}) = \sum_{c =1}^{N} Q(C=c) Q^{(c)}(X_{1:\ell} ) 
\end{align}
over length-$\ell$ sequences,
where $C$ is the random variable for the ergodic components,
and $Q^{(c)}(X_{1:\ell} ) := Q(X_{1:\ell} | C = c )$.

When a 
network is 
trained well on the nonergodic data source, and then
tested on a single ergodic component $c$, 
the expected in-context cross-entropy 
loss,
\begin{align}
\braket{\mathcal{L}_N^{(\ell)}}_{Q^{(c)}(X_{1:\ell+1})} 
&= 
- \sum_{x \in \mathcal{X}} \sum_{w \in \mathcal{X}^\ell} Q^{(c)}(wx) 
\log \biggl[\frac{\sum_{c'=1}^N Q(c') Q^{(c')}(wx)}{\sum_{c'=1}^N Q(c') Q^{(c')}(w)} \biggr]
\label{eq:InContextLossOnSingleComponent}
~,
\end{align}
depends on the number $N$ of ergodic components in the training data, as well as the structure of these other components.
Averaging this in-context loss over all components yields the myopic entropy rate $\braket{\braket{\mathcal{L}_N^{(\ell)}}_{Q^{(C)}(X_{1:\ell+1})}}_{Q(C)} = h_\ell^Q$.
When evaluating a particular component,
 the in-context loss of Eq.~\eqref{eq:InContextLossOnSingleComponent} reveals significantly more structure.

To isolate the key implications of nonergodicity, 
we train transformers on a minimal model of 
nonergodicity: sequence data generated by $N$ biased coins.
This provide a minimal model for power-law decay of in-context loss, which has been observed in scaling laws of transformers trained on natural language~\cite[Fig.~20]{Kaplan20_Scaling}. 
Our theoretical framework
allows us to easily anticipate the key features of in-context learning in this scenario.

The tokens within each sequence report the outcome of coin flips from a single coin,
but for each sequence the coin is drawn independently from a bag of biased coins with $N$ different biases,
where the $n^\text{th}$ coin comes up heads with probability $p_n$.
At context position $\ell$, there are exactly $\ell$ predictive states $(\predst_{k,\ell})_{k=1}^\ell$ that an observer could be in, where the $k^\text{th}$ predictive state  at the $\ell^\text{th}$ context position $\predst_{k,\ell}$ is the equivalence class of length-$\ell$ histories that have $k$ heads.
Notably, the MSP has the same topological structure as for two biased coins shown in the bottom row of Fig.~\ref{fig:MyopicEntropy2}, although the transition probabilities among predictive states depends on the number of ergodic components.


If a neural network minimizes expected loss on a nonergodic dataset of $N$ biased coins,
Appendix \ref{sec:Nonergodic_app} shows that
the expected in-context loss 
on sequences from coin $c$
will be 
\begin{align}
\braket{\mathcal{L}_N^{(\ell)}}_{\! Q^{(c)} \!(X_{1:\ell+1})} \!
&= 
-\! \sum_{k=0}^\ell 
\sum_{x \in \mathcal{X}}
\! Q^{(c)} \! (x)
{\ell \choose k}
p_c^{k} (1-p_c)^{\ell-k}
\log \Bigl[\tfrac{\sum_{c'=1}^N Q(c') Q^{(c')}(x) p_{c'}^{k} (1-p_{c'})^{\ell-k}}{\sum_{c'=1}^N Q(c') p_{c'}^{k} (1-p_{c'})^{\ell-k}} \Bigr] .
\end{align}
As an explicit example, 
we choose odd $N$ coins of bias evenly spaced on the interval $[0,1]$ such that
the $n^\text{th}$ coin comes up heads 
with probability $p_n = n/(N+1)$.

\begin{figure}
\centering
    \includegraphics[width=1.03\linewidth, trim={60 0 0 0}, clip]{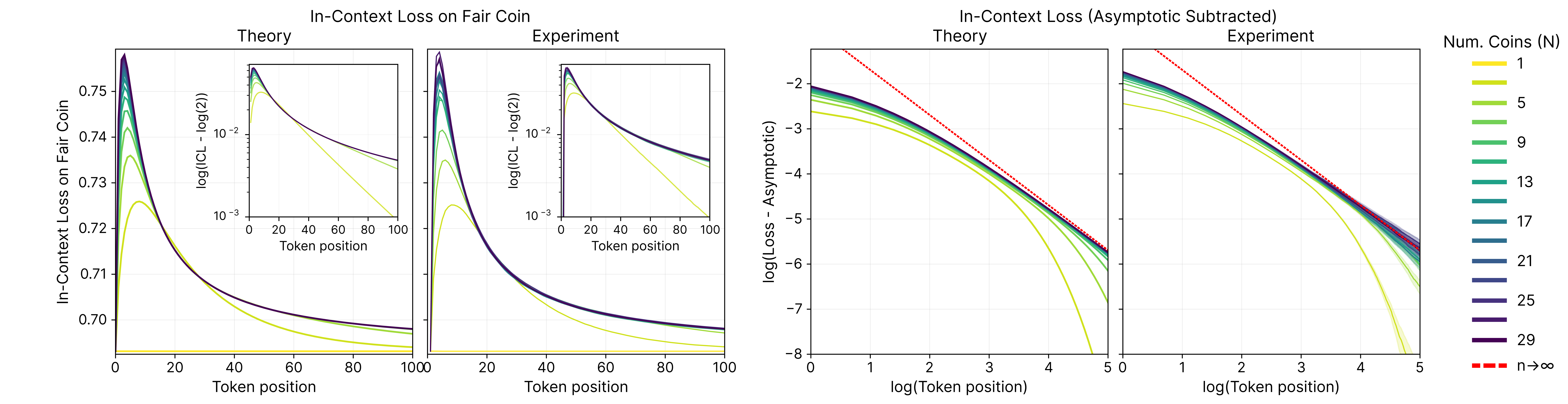}

	\caption{Experimental validation of in-context loss scaling behavior given by our framework. Left two panels:
    Expected in-context loss when presented with a fair coin, if the network is trained on $N$ coins of bias $\{ n/(N+1) \}_{n=1}^N$, for both theory and experiment.  Asymptotically, this uncertainty decays exponentially for small $N$ (indicated by the eventual straight line in the log-linear inset for $N=3$), but appears to decay significantly more slowly for a larger number of components.
    Right two panels: In-context loss (averaged over all coins) scales as a power-law with increasing context as the number of ergodic components grows large.  This 
    power-law decay of net loss is apparent from the straight line that emerges on the log-log plots for large $N$ and large token positions. The dotted red line shows the theoretically predicted asymptotic slope for large $\ell$ as $N \to \infty$. All experimental curves are plotted with error bars of $\pm 1$ standard error of the mean.
	}
\label{fig:NonergodicCoins_InContextLoss}
\end{figure}	

Fig.~\ref{fig:NonergodicCoins_InContextLoss} (left two panels)
shows the expected in-context loss when a neural network pretrained on sequences from these $N$ coins is presented with a fair coin.
Initially, in-context loss increases: As a heads or tails is observed, it biases the model to infer that it is observing a biased coin.  After working through some spurious correlations, the network eventually recognizes the approximate fairness of the coin and loss goes down.  Ref.~\cite{Ortega19_Meta} referred to this in-context learning scenario as meta-learning, and it should now be clear that this is a necessary implication of pretraining on non-ergodic sequences.

As the number of coins grows large, 
the myopic entropy converges to a perfect power law.  As derived in Appendix \ref{app:PowerLaw} and illustrated in the right two panels of Fig.~\ref{fig:NonergodicCoins_InContextLoss}, we find:
\begin{align}
\lim_{N \to \infty} ( h_\ell^{Q_N} - h^{Q_N}) &\approx \tfrac{1}{2} \ell^{-1 }  & \text{for large } \ell ~,
\label{eq:myopic_power_law}
\end{align}
where $h^{Q_N} = \lim_{\ell \to \infty} h_\ell^{Q_N} $ is the Shannon entropy rate of the process (reported in nats).
Accordingly---regardless of how many tokens you've seen---if you see again as many tokens, then you can expect your remaining excess uncertainty to half.~\footnote{Note that this halving property is because of the negative exponent rather than the 1/2 in Eq.~\eqref{eq:myopic_power_law}.}









\section{Limitations and future work}

We have shown that we can anticipate precise quantitative aspects of in-context learning.
However, this work has focused on in-distribution inference.
Ref.~\cite{Chan22_Data} argues that 
transformers may significantly outperform other common neural architectures at in-context learning 
out of distribution. 
Indeed, mechanisms like 
induction heads suggest that models should be able to generalize aspects of 
in-context learning even to realizations of motifs like skip trigrams that are out of the support of the training data.
We plan to investigate this out-of-support generalization in future work.

While we have observed in-context loss nearly converge to the myopic entropy rate of the correlated input process over the course of training, 
we hope that future research builds on these findings to help anticipate 
the detailed dynamics of training and in-context loss prior to convergence.  


\section{Conclusion}

This work has investigated,
both mathematically and empirically,
the extent to which 
the correlational structure
of training data 
implies a necessary architecture-independent
contribution to in-context learning.
Our experiments---on synthetic training datasets of differing types of correlational structure---reproduce characteristic phenomena like phase transitions in training loss for induction head formation and power-law scaling of ICL. We furthermore showed that a model’s in-context performance on any task is mathematically coupled to the ensemble of other tasks seen in pretraining.

By providing an experimentally verified grounded theoretical framework
and an
architecture-independent benchmark for ICL, 
this sets the stage for future investigations
that may seek architecture-specific aspects of generalization that go beyond what can be predicted from the data's information-theoretic properties.
Architecture-specific mechanisms for implementing in-distribution reduction in myopic entropy should be analyzed for how these mechanisms would generalize when presented with out-of-distribution contexts.

Finally, we note that 
our general results were not specific to language, and are thus not restricted to LLMs.
Rather, we should anticipate very generally that sequence models across modalities---text, vision, haptic, auditory, etc.---pretrained on next-token prediction, will exhibit direct analogs of ICL.


\section*{Acknowledgments}

We are grateful for Astera Institute's support of this work.
We appreciate the insights and inspirations from many wonderful colleagues. 




\bibliographystyle{unsrt}

\bibliography{chaos,ref}


\appendix

\section{Information theoretic interpretation of in-context loss}

For stationary processes,
$h_{\ell+1}^Q \leq h_\ell^Q$, so \emph{the next-token entropy of predictions from 
	any model well pretrained on an approximately stationary process must go down as it sees more context}.

Moreover, if the process generating the sequences approximates a stationary stochastic process, then the minimized loss per token would asymptote to the \emph{Shannon entropy rate} $h^Q = \lim_{\ell \to \infty} h_\ell^Q $, while the excess entropy accumulated on top of that
coincides with
the mutual information between past and future.  I.e.,
\begin{align}
	\min_{\params}
	\braket{\mathcal{L}_{\params}}_{Q(X_{1:L})}  
	&= 
	L h^Q + \EE_L
	~,
\end{align}	
where the so-called \emph{predictive information}~\cite{Bialek99_Predictive}
\begin{align}
	\EE_L = \sum_{\ell = 1}^L \bigl( h_\ell^Q - h^Q \bigr) = \I[ X_{1:L} ; X_{L+1:\infty}]
\end{align}	
is the amount of information that the past $X_{1:L}$ contains about the future $X_{L+1:\infty}$, which limits to the
so-called \emph{excess entropy}~\cite{Crut01a, Gras86}
$\EE = \lim_{L \to \infty} \EE_L
	= \I[ \overleftarrow{X} ; \overrightarrow{X}]$,
which is the mutual information $\I[\cdot ; \cdot]$ between the infinite past $\overleftarrow{X}$ and infinite future $\overrightarrow{X}$.



\section{Sample convergence of myopic entropy estimates from trained model}

Let $N_\ell$ be the random variable for the next-token entropy at token-position $\ell$, induced by the distribution over length-$\ell$ contexts.

We know that $N_\ell \in [0, \, \log|\mathcal{X}|]$.
In fact, we have the full distribution for $N_\ell$ via the predictive metadynamic, and powers of its transition operator:
\begin{align}
\Pr(N_\ell = n) = \sum_{\predst \in \predSSet_\ell} \braket{\delta_{\predfn(\varnothing)} | W^\ell | \delta_{\predst}} \, \delta_{n, \Shannon(X | \predst)} ~,
\end{align}
where 
$\predSSet_\ell := \{ Q(\overrightarrow{X} | X_{1:\ell} = w) : w \in \mathcal{X}^\ell \}$.

The expectation value of $N_\ell$ is 
itself the myopic entropy $h_\ell$ since 
\begin{align}
\braket{N_\ell} = \sum_{w \in \mathcal{X}^\ell} Q(w) \Shannon[X_{\ell+1} | X_{1:\ell} = w] = \Shannon[X_{\ell+1} | X_{1:\ell}] = h_\ell ~.
\end{align}

By the central limit theorem, 
if we estimate the myopic entropy via the conditional next-token entropy from a large number $M$ of sequences,
then the empirical estimate of myopic entropy $\hat{h}_\ell^{(M)} = \tfrac{1}{M} \sum_{m=1}^M n_\ell^{(m)}$ (i.e., the sample mean of $N_\ell$)
converges in distribution to the normal distribution
$\hat{h}_\ell^{(M)} \stackrel{d}{\to} \mathcal{N}(h_\ell, \sigma_M^2)$ with mean $h_\ell$ and variance $\sigma_M^2$,
where 
$\sigma_M^2 = \frac{\sigma^2}{M}$
and 
$\sigma^2 = \braket{N_\ell^2} - \braket{N_\ell}^2 \leq (\log| \mathcal{X}|)^2$.

\section{In-context Learning Mr. Durdsley/Wonka Example with RNN}

\begin{figure}[h!]
    \centering
    \includegraphics[width=1\linewidth]{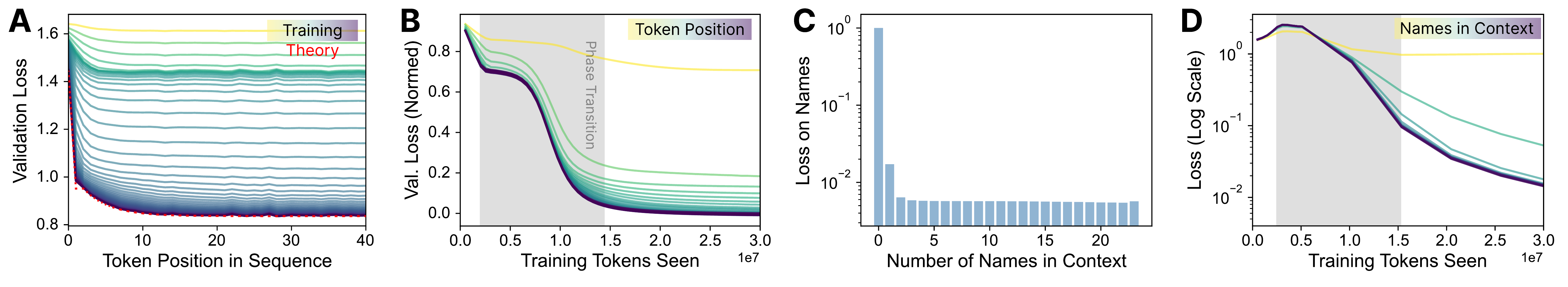}
    \caption{Same as Fig.~\ref{fig:InductionHeads}, but trained on a GRU RNN. Interestingly, the same phase transition associated with induction heads is seen in this non-transformer architecture, although the mechanism for in-context learning must be different because of different architectural constraints.}
    \label{fig:app_WonkaDursley}
\end{figure}

\begin{figure}[h!]
    \centering
    \includegraphics[width=0.9\linewidth]{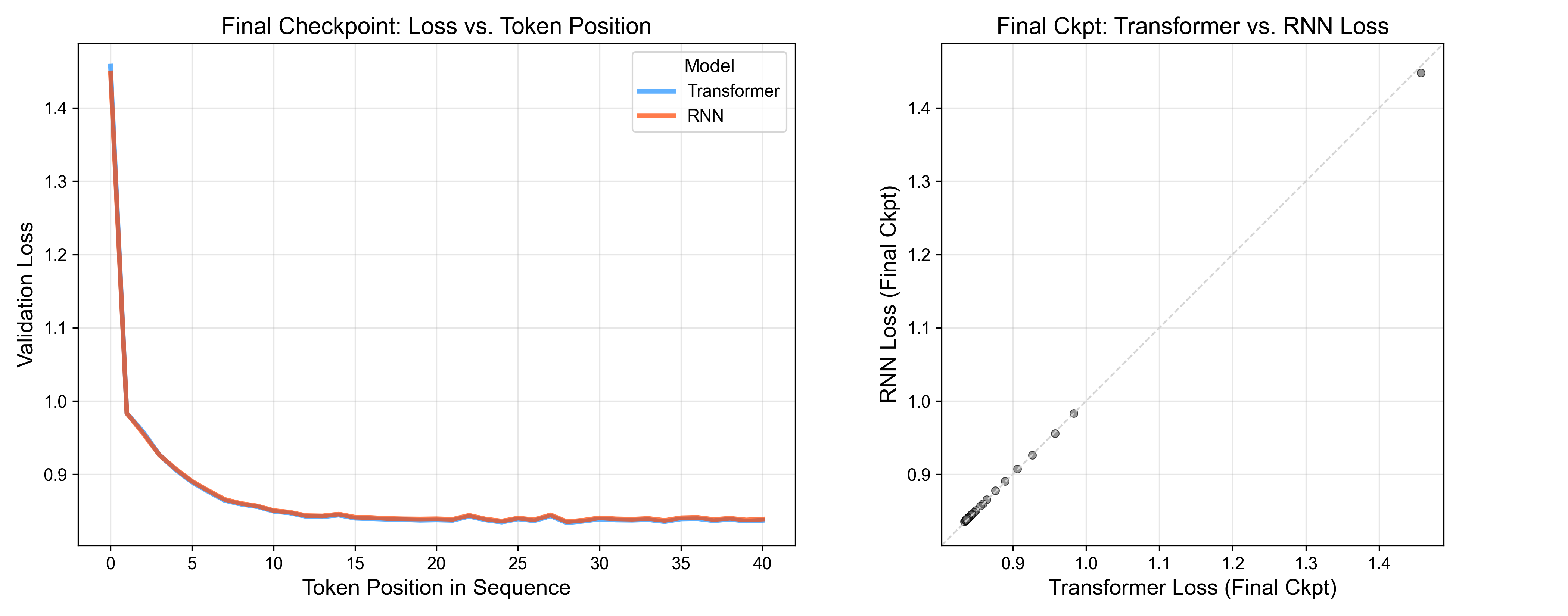}
    \caption{Comparison of Final Checkpoint In-Context Loss for RNN and Transformer Models on the Wonka--Dursley Process. (Left) Validation loss as a function of token position for the final training checkpoint of an RNN (red) and a Transformer (blue) on the Wonka--Dursley task. Both architectures achieve very similar in-context loss profiles, showing a reduction in loss (uncertainty) as more tokens in the sequence are processed. (Right) Scatter plot comparing the final checkpoint loss of the Transformer (x-axis) versus the RNN (y-axis) for each token position. Points falling along the y=x diagonal (dashed line) indicate equivalent performance at that token position. The close alignment suggests both architectures converge to a similar level of in-context learning performance on this task, despite their architectural differences.}
    \label{fig:app_Wonka_comp}
\end{figure}

The training curves for RNNs (Fig.~\ref{fig:app_WonkaDursley}) have some differences from those of transformers trained on the same process (Fig.~\ref{fig:InductionHeads}), but both exhibit signatures of a phase transition during training.  Nevertheless, RNNs and transformers achieve the same in-context loss on in-distribution tests (Fig.~\ref{fig:app_Wonka_comp}). Since RNNs cannot implement induction heads---they don't have the architectural wherewithal---future work will investigate their distinct methods for implementing this in-context reduction in uncertainty, and the implications for generalization for the two architectures on out-of-distribution sequences.

\section{Which OOD and OOS generalizations are consistent with low loss?}

If a neural network has achieved low loss on the training set of sequences of correlated random variables sampled from some process $Q$,
then what types of out-of-distribution (OOD) and out-of-support (OOS) generalizations are consistent with this low loss on the training set?

What is the best that the neural network could perform on some other process $Q'$?

We note that the neural network has learned what to do on all histories in the support of $Q$, but we do not a priori know how the network will act when conditioned on contexts outside of the support of $Q$.  
\begin{align}
\braket{\mathcal{L}_{\params^*}}_{Q'(X_{1:L})} 
&=
\sum_{\ell = 1}^L - \braket{ \log \Pr_{\params^*} (X_\ell | X_{1:\ell-1})}_{Q'(X_{1:\ell})} \\
& \geq \sum_{\ell=1}^L h_\ell^{Q'} ~.
\end{align}
If $\Pr_{\params}(X_{\ell+1} | X_{1:\ell} = w) = Q'(X_{\ell+1} | X_{1:\ell} = w)$ for all $w$ outside of $Q$'s support, then we obtain a best-case loss equal to the myopic entropy on this held out set.  However networks generalize, they can never achieve in-context loss below this fundamental limit.

\section{Nonergodic loss}
\label{sec:Nonergodic_app}

Here we provide a more detailed derivation of the equations for expected in-context loss when neural networks are trained on nonergodic processes.  This is the practically very relevant case where the statistics within a typical sequence are not representative of the statistics over all sequences.

Any nonergodic process with $N$ ergodic components
has the joint distribution 
\begin{align}
Q_N(X_{1:\ell}) = \sum_{c =1}^{N} Q(C=c) Q^{(c)}(X_{1:\ell} ) 
\end{align}
over length-$\ell$ sequences,
where $C$ is the random variable for the ergodic components,
and $Q^{(c)}(X_{1:\ell} ) := Q(X_{1:\ell} | C = c )$.

When a 
network is 
trained well on the nonergodic data source, and then
tested on a single ergodic component $c$, 
the expected in-context cross-entropy 
loss,
\begin{align}
\braket{\mathcal{L}_N^{(\ell)}}_{Q^{(c)}(X_{1:\ell+1})} &:= 
\braket{\mathcal{L}_N(X_{\ell+1} |X_{1:\ell})}_{Q^{(c)}(X_{1:\ell+1})} \\
&=
- \sum_{x \in \mathcal{X}} \sum_{w \in \mathcal{X}^\ell} Q^{(c)}(wx) 
\log Q_N(x|w) \\
&= 
- \sum_{x \in \mathcal{X}} \sum_{w \in \mathcal{X}^\ell} Q^{(c)}(wx) 
\log \biggl[\frac{\sum_{c'=1}^N Q(c') Q^{(c')}(wx)}{\sum_{c'=1}^N Q(c') Q^{(c')}(w)} \biggr]
~,
\end{align}
depends on the number $N$ of ergodic components in the training data, as well as the structure of these other components.
Averaging this in-context loss over all components yields the myopic entropy rate $\braket{\braket{\mathcal{L}_N(X_{\ell+1} |X_{1:\ell})}_{Q^{(C)}(X_{1:\ell+1})}}_{Q(C)} = h_\ell^Q$.

Let's now consider the useful minimal model of nonergodic sequence data generated by $N$ biased coin.
The tokens within each sequence report the outcome of coin flips from a single coin,
but for each sequence the coin is drawn independently from a bag of biased coins with $N$ different biases.
At context position $\ell$, there are exactly $\ell$ predictive states $(\predst_{k,\ell})_{k=1}^\ell$ that an observer could be in, where the $k^\text{th}$ predictive state  at the $\ell^\text{th}$ context position $\predst_{k,\ell}$ is the equivalence class of length-$\ell$ histories that have $k$ heads.
Notably, the MSP has the same topological structure as for two biased coins, although the transition probabilities among predictive states depends on the number of ergodic components.

%
If each ergodic component is a biased coin, then
$Q^{(c)}(w) = p_c^{k} (1-p_c)^{\ell-k}$ if 
$w \in \predst_{k,\ell}$
and thus
$Q^{(c)}(X_{1:\ell} \in \predst_{k,\ell}) = {\ell \choose k} p_c^{k} (1-p_c)^{\ell-k}$,
where 
the $n^\text{th}$ coin comes up heads with probability $p_n$.

If a neural network minimizes expected loss on a nonergodic dataset of $N$ biased coins,
the expected in-context loss 
on sequences from coin $c$
will be 
\begin{align}
\braket{\mathcal{L}_N^{(\ell)}}_{Q^{(c)}(X_{1:\ell+1})}
&= 
- \!
\sum_{k =0}^{\ell}
\sum_{x \in \mathcal{X}}
Q^{(c)}(x | \predst_{k,\ell}) 
Q^{(c)}( \predst_{k,\ell})
\log \biggl[\tfrac{\sum_{c'=1}^N Q(c') Q^{(c')}(x| \predst_{k,\ell}) p_{c'}^{k} (1-p_{c'})^{\ell-k}}{\sum_{c'=1}^N Q(c') p_{c'}^{k} (1-p_{c'})^{\ell-k}} \biggr]  \\
&= 
- \!
\sum_{k=0}^\ell 
\sum_{x \in \mathcal{X}}
Q^{(c)}(x)
{\ell \choose k}
p_c^{k} (1-p_c)^{\ell-k}
\log \biggl[\tfrac{\sum_{c'=1}^N Q(c') Q^{(c')}(x) p_{c'}^{k} (1-p_{c'})^{\ell-k}}{\sum_{c'=1}^N Q(c') p_{c'}^{k} (1-p_{c'})^{\ell-k}} \biggr] ,
\label{eq:ICL_NBC_tested_on_one_component}
\end{align}
where we note that, for large $\ell$,
the binomial probability
${\ell \choose k}
p_c^{k} (1-p_c)^{\ell-k}$ can be approximated by the normal distribution
$\mathcal{N} \bigl( p_c \ell , \, p_c (1-p_c) \ell  \bigr) = \tfrac{1}{\sqrt{2 \pi p_c(1-p_c) \ell}} e^{\tfrac{-(k - p_c \ell)^2}{2 p_c (1-p_c) \ell}}$ with mean $p_c \ell$ and variance
$p_c (1-p_c) \ell$.

As an explicit example, 
we choose odd $N$ coins of bias evenly spaced on the interval $[0,1]$ such that
the $n^\text{th}$ coin comes up heads 
with probability $p_n = n/(N+1)$,
and each coin is drawn with equal probability to produce each sequence.
I.e., $Q(c') = 1/N$ for all components $c'$, 
and $Q^{(n)}(X) = (p_{n}, 1-p_{n}) = (\tfrac{n}{N+1}, \tfrac{N-n+1}{N+1})$.
Eq.~\eqref{eq:ICL_NBC_tested_on_one_component}
then further simplifies to 
\begin{align}
\braket{\mathcal{L}_N^{(\ell)}}_{Q^{(c)}(X_{1:\ell+1})}
&= 
-\sum_{k=0}^\ell 
{\ell \choose k}
p_c^{k} (1-p_c)^{\ell-k}
\Biggl\{
p_c
\log \biggl[\frac{\sum_{c'=1}^N p_{c'}^{k+1} (1-p_{c'})^{\ell-k}}{\sum_{c'=1}^N p_{c'}^{k} (1-p_{c'})^{\ell-k}} \biggr] \nonumber \\
& \qquad \qquad \qquad \qquad \qquad
+
(1-p_c)
\log \biggl[\frac{\sum_{c'=1}^N p_{c'}^{k} (1-p_{c'})^{\ell+1-k}}{\sum_{c'=1}^N p_{c'}^{k} (1-p_{c'})^{\ell-k}} \biggr]
\Biggr\}
~.
\label{eq:app_ICL_NBC_tested_on_one_component}
\end{align}
When we evaluate the in-context loss on the fair coin $c = (N+1)/2$, this expression simplifies to 
\begin{align}
\braket{\mathcal{L}_N^{(\ell)}}_{\! Q^{((N+1) /2)} \! (X_{1:\ell+1})} \! &=
- \! \sum_{k=0}^\ell {\ell \choose k} 2^{-(\ell+1)} \log \biggl( \tfrac{ \bigl[ \sum_{c'=1}^N p_{c'}^{k+1}(1-p_{c'})^{\ell-k} \bigr] 
\bigl[ \sum_{c'=1}^N p_{c'}^{k}(1-p_{c'})^{\ell+1-k} \bigr] }{ \bigl[ \sum_{c'=1}^N p_{c'}^{k}(1-p_{c'})^{\ell-k} \bigr]^2 } \biggr) .
\end{align}
If instead we're interested in the expected loss over all contexts for a well trained network,
we obtain the myopic entropy rate
\begin{align}
h_\ell^Q
&= 
-\tfrac{1}{N} \sum_{c=1}^N \sum_{k=0}^\ell 
{\ell \choose k}
p_c^{k} (1-p_c)^{\ell-k}
\Biggl\{
p_c
\log \biggl[\frac{\sum_{c'=1}^N p_{c'}^{k+1} (1-p_{c'})^{\ell-k}}{\sum_{c'=1}^N p_{c'}^{k} (1-p_{c'})^{\ell-k}} \biggr] 
\nonumber \\
& \qquad \qquad \qquad \qquad \qquad
\qquad \qquad
+
(1-p_c)
\log \biggl[\frac{\sum_{c'=1}^N p_{c'}^{k} (1-p_{c'})^{\ell+1-k}}{\sum_{c'=1}^N p_{c'}^{k} (1-p_{c'})^{\ell-k}} \biggr]
\Biggr\}
~.
\label{eq:h_ell_NBC}
\end{align}

\subsection{Power law loss convergence from many ergodic components}
\label{app:PowerLaw}

In the limit of many ergodic components, $N\to\infty$, we can exactly evaluate the in-context loss in closed form, and show that it scales as a power law for large $\ell$.

To achieve this, we leverage the fact that~\footnote{This may be easiest to understand in relation to the Beta distribution, which is a special case of the Dirichlet distribution---each being a parametrized distribution over distributions.} 
\begin{align}
g(a,b) :=
\lim_{N\to\infty} \tfrac{1}{N} \sum_{n=1}^N p_n^a (1-p_n)^b 
&= \int_{p=0}^1 dp \, p^a (1-p)^b 
= \frac{a! \, b!}{(a+b+1)!} & \text{for } a, b \in \mathbb{Z}_{\geq 0}~.
\end{align}
Eqs.~\eqref{eq:app_ICL_NBC_tested_on_one_component} and \eqref{eq:h_ell_NBC}
each contain several instances of 
$g(a,b)$ with non-negative integers $a$ and $b$.

In this limit of many ergodic components, test loss on a single component becomes
\begin{align}
\braket{\mathcal{L}_\infty^{(\ell)}}_{Q^{(c)}(X_{1:\ell+1})} &= \lim_{N \to \infty}
\braket{\mathcal{L}_N^{(\ell)}}_{Q^{(c)}(X_{1:\ell+1})} \\
&= 
-\sum_{k=0}^\ell 
{\ell \choose k}
p_c^{k} (1-p_c)^{\ell-k}
\Biggl\{
p_c
\log \biggl[\frac{g(k+1,\ell-k) }{g(k,\ell-k) } \biggr] 
\nonumber \\
& \qquad \qquad \qquad \qquad \qquad
+
(1-p_c)
\log \biggl[\frac{g(k,\ell+1-k) }{g(k,\ell-k) } \biggr]
\Biggr\} 
\\
%
&= 
-\sum_{k=0}^\ell 
{\ell \choose k}
p_c^{k} (1-p_c)^{\ell-k}
\Biggl\{
p_c
\log \biggl[\frac{(k+1)! \, (\ell+1)! }{k! \, (\ell+2)! } \biggr] 
\nonumber \\
& \qquad \qquad \qquad \qquad \qquad
+
(1-p_c)
\log \biggl[\frac{ (\ell+1-k)! \, (\ell+1)! }{(\ell-k)! \, (\ell+2)!} \biggr]
\Biggr\} \\
&= 
-\sum_{k=0}^\ell 
{\ell \choose k}
p_c^{k} (1-p_c)^{\ell-k}
\Biggl[
p_c
\log \biggl(\frac{k+1 }{\ell+2} \biggr) 
+
(1-p_c)
\log \biggl( \frac{ \ell+1-k  }{ \ell+2} \biggr)
\Biggr] 
~.
\label{eq:app_ICL_NBC_tested_on_one_component_Ninf}
\end{align}

Expected loss on a fair coin (where $p_c = 1/2$) is then
\begin{align}
\braket{\mathcal{L}_\infty^{(\ell)}}_{Q \sim \text{Fair}} 
&= 
-\sum_{k=0}^\ell 
{\ell \choose k}
2^{-(\ell+1)}
\log \biggl[\frac{(k+1)(\ell+1-k) }{(\ell+2)^2 } \biggr] 
~.
\end{align}


Averaging Eq.~\eqref{eq:app_ICL_NBC_tested_on_one_component_Ninf} over all components,
the myopic entropy rate in the limit of many components becomes
\begin{align}
h_\ell^{Q_\infty} &= \lim_{N\to\infty} h_\ell^{Q_N} \\
&= 
-\lim_{N \to \infty} 
\tfrac{1}{N}
\sum_{c=1}^N
\sum_{k=0}^\ell 
{\ell \choose k}
p_c^{k} (1-p_c)^{\ell-k}
\Biggl[
p_c
\log \biggl(\frac{k+1 }{\ell+2} \biggr) 
+
(1-p_c)
\log \biggl( \frac{ \ell+1-k  }{ \ell+2} \biggr)
\Biggr]  
\\ 
&= 
-\lim_{N \to \infty} 
\tfrac{1}{N}
\sum_{k=0}^\ell 
{\ell \choose k}
\Biggl\{
\biggl[
\sum_{c=1}^N
p_c^{k+1} (1-p_c)^{\ell-k}
\biggr]
\log \biggl(\frac{k+1 }{\ell+2} \biggr) 
\nonumber \\
& \qquad \qquad \qquad \qquad \quad \;
+
\biggl[
\sum_{c=1}^N
p_c^{k} (1-p_c)^{\ell+1-k}
\biggr]
\log \biggl( \frac{ \ell+1-k  }{ \ell+2} \biggr)
\Biggr\} 
\\
& = 
\sum_{k=0}^\ell 
{\ell \choose k}
\Biggl[
g(k+1, \ell-k)
\log \biggl(\frac{k+1 }{\ell+2} \biggr) 
+
g(k, \ell+1-k)
\log \biggl(\frac{\ell+1-k}{\ell+2} \biggr) 
\Biggr]
\\
& = 
-
\sum_{k=0}^\ell 
\frac{\ell!}{ k! \, (\ell-k)!}
\Biggl[
\frac{(k+1)! \, (\ell-k)!}{(\ell+2)!}
\log \biggl(\frac{k+1 }{\ell+2} \biggr) 
+
\frac{k! (\ell+1-k)!}{(\ell+2)!}
\log \biggl(\frac{\ell+1-k}{\ell+2} \biggr) 
\Biggr]
\\
&= \frac{1}{\ell+1} 
\sum_{k=0}^\ell 
\Biggl[ 
-
\frac{(k+1) }{(\ell+2)}
\log \biggl(\frac{k+1 }{\ell+2} \biggr) 
-
\frac{(\ell+1-k)}{(\ell+2)}
\log \biggl(\frac{\ell+1-k}{\ell+2} \biggr) 
\Biggr] \\
& = 
\frac{1}{\ell+1} 
\sum_{k=0}^\ell 
B_2
\Bigl(\frac{k+1 }{\ell+2} \Bigr) 
~,
\label{eq:app_h_ell_Ninf}
\end{align}
where $B_2(q) = -q \log(q) - (1-q) \log (1-q)$ is the binary entropy function.

The asymptotic entropy rate is
\begin{align}
h^{Q_\infty} = 
\lim_{\ell \to \infty}
\frac{1}{\ell+1}
\sum_{k=0}^\ell 
B_2
\Bigl(\frac{k+1 }{\ell+2} \Bigr) 
=
\int_{p=0}^1 B_2(p) \, dp
= \tfrac{1}{2} \text{ nats / token}
~.
\label{eq:app_h_Ninf}
\end{align}

Meanwhile, for large but finite $\ell$, the sum in Eq.~\eqref{eq:app_h_ell_Ninf} approximates the constant definite integral of the binary entropy function, such that:
\begin{align}
h_\ell^{Q_\infty} &= 
\frac{\ell+2}{\ell+1} \frac{1}{\ell+2}
\sum_{k=0}^\ell 
B_2
\Bigl(\frac{k+1 }{\ell+2} \Bigr) \\
& \approx
\frac{\ell+2}{\ell+1} 
\int_{p=0}^1 B_2(p) \, dp & \text{for large } \ell \\
& = 
\tfrac{1}{2} (1 + \tfrac{1}{\ell+1})
~,
\end{align}
from which it follows that the myopic entropy rate converges to its asymptotic value as a power law, with exponent $\alpha=-1$, for large $\ell$:
\begin{align}
h_\ell^{Q_\infty} - h^{Q_\infty}
& \approx \tfrac{1}{2} \tfrac{1}{\ell+1} & \text{for large } \ell \\
&= \tfrac{1}{2} \sum_{n=1}^\infty (-1)^{n+1} \ell^{-n} \\
&= \tfrac{1}{2} \ell^{-1} + \mathcal{O}(\ell^{-2}) \\
&\approx \tfrac{1}{2} \ell^{-1}
& \text{for large } \ell ~.
\end{align}

\begin{figure}
\centering
    \includegraphics[width=0.75\linewidth]{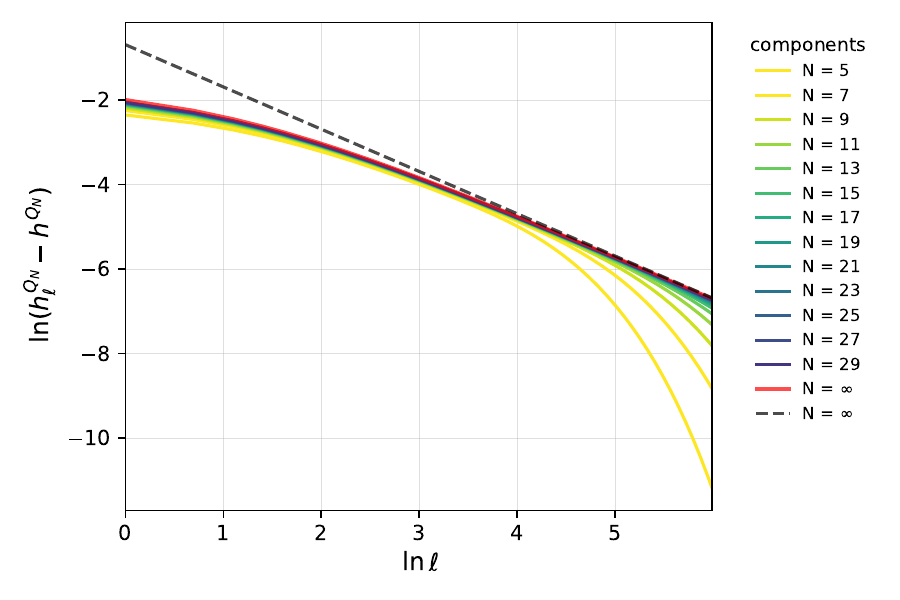}
	
	\caption{Expected in-context loss (averaged over all coins) scales as a power-law with increasing context as the number of ergodic components grows large.  
    The dashed gray line from Eq.~\eqref{eq:app_asymptotic_myopic_entropy_scaling} shows that the 
    myopic entropy rate decays to its asymptotic value as $\ell^{-1}$ for large $\ell$, in the limit of many ergodic components ($N \to \infty$).
	}
\label{fig:app_NonergodicCoins_InContextLoss}
\end{figure}

This power-law behavior is verified by linearity in the log-log plot of the expected loss for transformers trained on many coins in Fig.~\ref{fig:NonergodicCoins_InContextLoss} of the main text, where
\begin{align}
\log (h_\ell^{Q_\infty} - h^{Q_\infty}) 
&\approx  -\log(2) - \log(\ell) & \text{for large } \ell ~.
\label{eq:app_asymptotic_myopic_entropy_scaling}
\end{align}
Eq.~\eqref{eq:app_asymptotic_myopic_entropy_scaling} is shown as the dashed line in Fig.~\ref{fig:app_NonergodicCoins_InContextLoss}.

\section{ML Architectures and experimental details}

For our experiments, we trained both transformers and RNNs.  

We found that the use of a beginning-of-sequence (BOS) token was important for the transformer's ability to minimize loss at the first few token positions.
We attribute this to the transformers' use of the BOS token as an attention sink and also its ability to assign a unique OV vector to the BOS token, which
eases the tension between optimal transient and long-term behavior.

All transformer models were based on Penzai's llama-like transformer.  Figure \ref{fig:NonergodicCoins_InContextLoss} training runs used a 5-layer model with 8 heads, embedding dim of 256, feed-forward hidden dim of 512, and projection dim of 32, with swiglu as the non-linear activation function, rope wavelength of 50, post-attention and post-ffw normalization layers.  The embedding and de-embedding matrices were not tied.  We used AdamW optimizer with zero weight decay and no Nesterov momentum, and the ``inverse square root'' learning rate schedule as described in 
Attention Is All You Need 
\cite[Eq.~(3)]{Vaswani17_Attention}
(\url{https://arxiv.org/abs/1706.03762}), with 50k warm-up steps and $d_\text{model}$ = 256.  Training data were sampled from each respective process with a context length of 100 (including the BOS token) and batch size of 4000.  Models were trained with traditional cross-entropy loss between model probabilities and one-hot encoded symbols from the generative process.  We note that training was quite robust and that several other architectural variants also trained well.

Figure \ref{fig:MyopicEntropy1} training runs used a 2-layer model with 2 heads, a query-head multiplier of 2, embedding dim of 16, mlp dim 16, and projection dim 16, a geglu\_approx non-linear activation, rope wavelength 1000, and no normalization layers.  We used AdamW optimizer with zero weight decay and no Nesterov momentum, and a constant learning rate of 0.001 and context length of 100 (including BOS token) and a batch size of 10000.  We found that the fit to the theoretical cross-entropy curves of the $N$-coin process versus the fair coin process depended critically on the use of a BOS token.  Without it, both the 2-layer and 5-layer transformer variants failed to fit the first few context positions even approximately.  With the BOS token, other architectural and training details were not critical.

For both Figures \ref{fig:MyopicEntropy1} and \ref{fig:NonergodicCoins_InContextLoss}, reported in-context metrics are the average of the metric computed on each of 200k realizations generated from a target process.

All networks were trained on H100 GPUs on a cloud provider. Given the small model sizes and synthetic nature of the training data, most training runs completed in less than one hour on an H100.

\subsection{Wonka--Dursley Process}

A transformer and RNN were trained on the Wonka--Dursley process described in Section~\ref{sec:InductionHeads} using the training parameters~\footnote{The BOS token prepended to each sequence is not counted as part of the sequence length.}~\footnote{The default values for \protect\href{https://optax.readthedocs.io/en/latest/api/optimizers.html}{Optax's Adam} optimizer were used.} listed in Table~\ref{dursley-wonka-training-params}.

\begin{table}[]
  \caption{Wonka--Dursley Training Parameters}
  \label{dursley-wonka-training-params}
  \centering
  \begin{tabular}{ll}
    \toprule
    Name & value \\
    \midrule
    sequence length & $100$ \\
    batch size & $512$ \\
    training steps & $1000$ \\
    optimizer & Adam \\
    learning rate & $0.001$ \\
    \bottomrule
  \end{tabular}
\end{table}

\protect\href{https://penzai.readthedocs.io/en/latest/_autosummary/penzai.models.transformer.variants.llamalike_common.html}{Penzai's Llamalike common} transformer architecture was used with the parameters listed in Table~\ref{dursley-wonka-transformer-params} (additionally the default values from \protect\href{https://penzai.readthedocs.io/en/latest/_autosummary/leaf/penzai.models.transformer.variants.llamalike_common.LlamalikeTransformerConfig.html}{Penzai's LlamalikeTransformerConfig} were used).

The RNN with parameters listed in Table~\ref{dursley-wonka-rnn-params} was comprised of a sequence of \protect\href{https://docs.kidger.site/equinox/api/nn/rnn/#equinox.nn.GRUCell}{Equinox's GRUCells} and a final linear layer to produce the output logits.

\begin{table}
  \caption{Wonka--Dursley Transformer Parameters}
  \label{dursley-wonka-transformer-params}
  \centering
  \scalebox{0.9}{
  \begin{tabular}{lll}
    \toprule
    \multicolumn{2}{c}{Part}                   \\
    \cmidrule(r){1-2}
    Name & Description & value \\
    \midrule
    num\_kv\_heads & The number of key-value attention heads or head groups & $2$     \\
    query\_head\_multiplier & The number of query heads for each KV head & $2$      \\
    embedding\_dim & Dimension of the embedding vectors and residual stream & $16$  \\
    projection\_dim & Dimension of the query, key, and value projections & $16$ \\
    num\_decoder\_blocks & Number of transformer decoder blocks in the model & $2$ \\
    mlp\_variant & Gated linear unit variant for MLPs & geglu\_approx \\
    tie\_embedder\_and\_logits & Whether to tie the weights of the embedding and logit layers & false \\
    \bottomrule
  \end{tabular}
  }
\end{table}

\begin{table}
  \caption{Wonka--Dursley RNN Parameters}
  \label{dursley-wonka-rnn-params}
  \centering
  \begin{tabular}{lll}
    \toprule
    \multicolumn{2}{c}{Part}                   \\
    \cmidrule(r){1-2}
    Name & Description & value \\
    \midrule
    hidden\_size & The dimensionality of the hidden state passed along between time steps & $16$  \\
    num\_layers & The number of GRU Cell layers & $2$     \\
    \bottomrule
  \end{tabular}
\end{table}



\end{document}